\def\year{2020}\relax
\documentclass[letterpaper]{article} 
\usepackage{aaai20}  
\usepackage{times}  
\usepackage{helvet} 
\usepackage{courier}  
\usepackage[hyphens]{url}  
\usepackage{graphicx} 
\usepackage{graphicx}  
\newcommand{\samethanks}{\footnotemark[1]}
\usepackage{graphicx,amssymb,amsmath,xcolor,amsthm,algorithm,algorithmic,url,multirow,array,nicefrac}
\newcommand{\tabincell}[2]{\begin{tabular}{@{}#1@{}}#2\end{tabular}}

\def\mcL{\mathcal{L}}
\def\bbE{\mathbb{E}}
\DeclareRobustCommand\onedot{\futurelet\@let@token\@onedot}
\def\@onedot{\ifx\@let@token.\else.\null\fi\xspace}
\def\ie{\emph{i.e}\onedot}

\usepackage{placeins}

\usepackage{natbib}

\usepackage{graphics}
\usepackage{graphicx}
\usepackage{subcaption}
\usepackage{float}

\usepackage{cleveref}

\title{Universal Adversarial Training}

\author{Ali Shafahi\textsuperscript{\rm }\thanks{Equal contribution.}, Mahyar Najibi\samethanks{}, Zheng Xu\samethanks{}, John Dickerson, Larry S. Davis, Tom Goldstein\\  
\textsuperscript{\rm }Deparment of Computer Science\\ 
University of Maryland\\
College Park, Maryland 20742\\
\{ashafahi, najibi, xuzh, john, lsd,tomg\}@cs.umd.edu 
}
 
\begin{document}

\maketitle

\begin{abstract}
Standard adversarial attacks change the predicted class label of a selected image by adding specially tailored small perturbations to its pixels.  In contrast, a {\em universal} perturbation is an update that can be added to {\em any} image in a broad class of images, while still changing the predicted class label. 
We study the efficient generation of universal adversarial perturbations, and also efficient methods for hardening networks to these attacks. We propose a simple optimization-based universal attack that reduces the top-1 accuracy of various network architectures on ImageNet to less than 20\%, while learning the universal perturbation $13\times$ faster than the standard method. 

To defend against these perturbations, we propose {\em universal adversarial training}, which models the problem of robust classifier generation as a two-player min-max game, and produces robust models with only $2\times$ the cost of natural training. We also propose a simultaneous stochastic gradient method that is almost free of extra computation, which allows us to do universal adversarial training on ImageNet.  
\end{abstract}

\section{Introduction}
Deep neural networks (DNNs) are vulnerable to adversarial examples, in which small and often imperceptible perturbations change the class label of an image \citep{szegedy2013intriguing,goodfellow6572explaining,nguyen2015deep,papernot2016limitations}. Many works have shown that these vulnerabilities can be exploited by showing real world attacks on face detection \citep{sharif2016accessorize}, object detection \citep{wu2019making}, and copyright detection \citep{saadatpanah2019adversarial}.

Adversarial examples were originally formed by selecting a single example, and sneaking it into a different class using a small perturbation  \citep{carlini2017towards}. This is done most effectively using (potentially expensive) iterative optimization procedures \citep{dong2017boosting,madry2017towards,athalye2018obfuscated}.  

\begin{figure}
    \centering
    \includegraphics[width=\columnwidth]{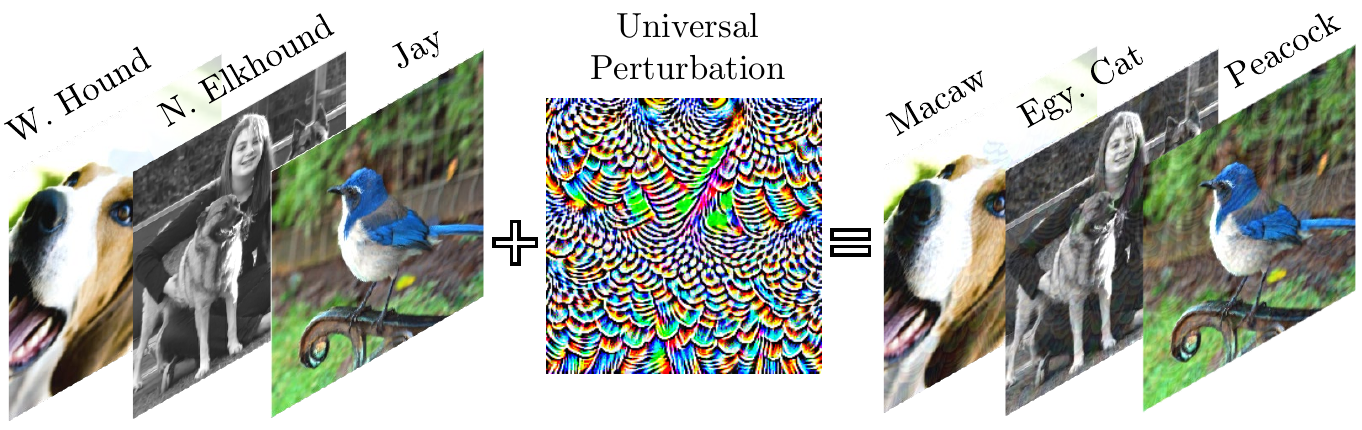}
    \caption{
    A universal perturbation made using a subset of ImageNet and the VGG-16 architecture. When added to the 
    validation images, their labels usually 
    change. The perturbation was generated 
    using the proposed \cref{alg:univ}. 
    Perturbation pixel values lie in 
    $[-10,10]$ 
    (i.e. 
    $\epsilon=10$). 
    }
    \label{fig:teaser}
\end{figure}

Different from per-instance perturbation attacks, 
\cite{moosavi2017universal,moosavi2017analysis} show there exists ``universal'' perturbations that can be added to any image to change its class label (\cref{fig:teaser}) with high probability.
Universal perturbations empower attackers who cannot generate per-instance adversarial examples on the go, or who want to change the identity of an object to be selected later in the field. 
Also, universal perturbations have good cross-model transferability, which facilitates black-box attacks. 

Among various methods for hardening networks to per-instance attacks, {\em adversarial training} \citep{madry2017towards} is known to dramatically increase robustness \cite{athalye2018obfuscated}. In this process, adversarial examples are produced for each mini-batch during training, and injected into the training data.  
While effective at increasing robustness against small perturbations, it is not effective for larger perturbations which are often the case for universal perturbations. Also, the high cost of this process precludes its use on large and complex datasets. 

\paragraph{Contributions}
This paper studies effective methods for producing and deflecting universal adversarial attacks.
First, we pose the creation of universal perturbations as an optimization problem that can be effectively solved by stochastic gradient methods.  This method dramatically reduces the time needed to produce attacks as compared to \cite{moosavi2017universal}. 
The efficiency of this formulation empowers us to consider universal adversarial training.  We formulate the adversarial training problem as a min-max optimization where the minimization is over the network parameters and the maximization is over the universal perturbation. This problem can be solved quickly using alternating stochastic gradient methods with no inner loops, making it far more efficient than per-instance adversarial training with a strong adversary (which requires a PGD inner loop to generate adversarial perturbations).
Prior to our work, it was argued that adversarial training on universal perturbations is infeasible because the inner optimization requires the generation of a universal perturbation from scratch using many expensive iterations \citep{perolat2018playing}. 
We further improve the defense efficiency by providing a ``low-cost'' algorithm for defending against universal perturbations. Through experiments on CIFAR-10 and ImageNet, we show that this ``low-cost'' method works well in practice. Note, the approach first introduced here has been expanded on to create  a range of ``free'' adversarial training strategies with the same cost as standard training \citep{shafahi2019free}.

\section{Related work}
We briefly review {\em per-instance} perturbation attack techniques that are closely related to our paper and can be used during the universal perturbation update step of universal adversarial training. The Fast Gradient Sign Method (FGSM) \citep{goodfellow6572explaining} is one of the most popular one-step gradient-based approaches for $\ell_\infty$-bounded attacks. FGSM applies one step of gradient ascent in the direction of the sign of the gradient of the loss function with respect to the input image.  When a model is FGSM adversarially trained, the gradient of the loss function may be very small near unmodified images.  In this case, the R-FGSM method remains effective by first using a random perturbation to step off the image manifold, and then applying FGSM \citep{tramer2017ensemble}.  Projected Gradient Descent (PGD) \citep{madry2017towards} iteratively applies FGSM multiple times, and is one of the strongest per-instance attacks \citep{athalye2018obfuscated}. The PGD version of \cite{madry2017towards} applies an initial random perturbation before multiple steps of gradient ascent and projection of perturbation onto the norm-ball of interest, and in a standard projected gradient method \citep{goldstein2014field}.
Finally, DeepFool \citep{moosavi2016deepfool} is an iterative method based on a linear approximation of the training loss objective.  

{\em Adversarial training}, in which adversarial examples are injected into the dataset during training, is an effective method to learn a robust model resistant to per-instance attacks \citep{madry2017towards,huang2015learning,shaham2015understanding,sinha2018certifying}. 
Robust models adversarially trained with FGSM can resist FGSM attacks \citep{kurakin2016adversarial}, but can be vulnerable to PGD attacks \citep{madry2017towards}.
\cite{madry2017towards} suggest strong attacks are important, and they use the iterative PGD method in the inner loop for generating adversarial examples when optimizing the min-max problem. PGD adversarial training is effective but time-consuming when the perturbation is small.  The cost of the inner PGD loop is high, although this can sometimes be replaced with neural models for attack generation \citep{baluja2017adversarial,poursaeed2017generative,xiao2018generating}. 
These robust models are adversarially trained to fend off per-instance perturbations and have not been designed for, or tested against, universal perturbations.  

Unlike per-instance perturbations, {\em universal} perturbations can be directly added to any test image to fool the classifier. 
In \cite{moosavi2017universal}, universal perturbations for image classification are generated by iteratively optimizing the per-instance adversarial loss for training samples using DeepFool. In addition to classification tasks,
universal perturbations are also shown to exist for semantic segmentation \citep{metzen2017universal}.
Robust universal adversarial examples are generated as a universal targeted adversarial patch in \cite{brown2017adversarial}. They are targeted since they cause misclassification of the images to a given target class.
\cite{moosavi2017analysis} prove the existence of small universal perturbations under certain curvature conditions of decision boundaries. 
Data-independent universal perturbations are also shown to exist and can be generated by maximizing spurious activations at each layer. These universal perturbations are slightly weaker than the data dependent approaches \citep{mopuri2017fast,mopuri2018generalizable}. As a variant of universal perturbation, unconditional generators are trained to create perturbations from random noises for attack \citep{reddy2018ask,reddy2018nag}. Universal perturbations are often larger than per-instance perturbation. For example on ImageNet, universal perturbations generated in prior works have $\ell_\infty$ perturbations of size $\epsilon=10$ while non-targeted per-instance perturbations as small as $\epsilon=2$ are often enough to considerably degrade the performance of conventionally trained classifiers possibly due to the fact that ImageNet is a complex dataset and is fundamentally susceptible to per-instance perturbations \citep{shafahi2018adversarial}. As a consequence, from a defense perspective, per-instance defenses on ImageNet focus on smaller perturbations compared to universal perturbations.

There has been very little work on defending against universal attacks. 
To the best of our knowledge, the only dedicated study is by \cite{akhtar2017defense}, who propose 
a perturbation rectifying network that pre-processes input images to remove the universal perturbation. The rectifying network is trained on universal perturbations that are built for the downstream classifier. While other methods of data sanitization exist 
\citep{samangouei2018defense,meng2017magnet} , it has been shown (at least for per-instance adversarial examples) that this type of defense is easily subverted by an attacker who is aware that a defense network is being used \citep{carlini2017adversarial}.

Two recent preprints \citep{perolat2018playing,mummadi2018defending} model the problem of defending against universal perturbations as a two-player min-max game. However, unlike us, and similar to per-instance adversarial training, after each gradient descent iteration for updating the DNN parameters, they generate a universal adversarial example in an iterative fashion. Since the generation of universal adversarial perturbations can be very time-consuming, this makes their approach slow and prevents them from training the DNN parameters for many iterations.

\section{Optimization for universal perturbation} \label{sec:adv_gen_opt}
 Given a set of training samples 
  $X = \{ x_i, i=1,\ldots, N \}$ 
 and a network $f(w, \cdot)$ with frozen parameter $w$ that maps images onto labels,  \cite{moosavi2017universal} propose to find universal perturbations $\delta$ that satisfy,
 \begin{equation}
 \| \delta \|_p \leq \epsilon \text{ and } \text{Prob}(X, \delta) \geq 1 - \xi, \label{eq:cvpr}
\end{equation}  
$ \text{Prob}(X, \delta)$ represents the ``fooling ratio,'' which is the fraction of images $x$ whose perturbed class label $f(w,x+\delta)$ differs from the original label $f(w,x)$. The parameter $\epsilon$ controls the $\ell_p$ diameter of the bounded perturbation, and $\xi$ is a small tolerance hyperparameter.  Problem \eqref{eq:cvpr} is solved by the iterative method in \cref{alg:cvpr}.
This solver relies on an inner loop to apply DeepFool to each training instance, which makes the solver slow. Moreover, the outer loop of \cref{alg:cvpr} is not guaranteed to converge.
\begin{algorithm}[t]
	\caption{
	\small	Iterative solver for universal perturbations \citep{moosavi2017universal}
	}
	\label{alg:cvpr}
	\begin{algorithmic}
		\STATE Initialize $\delta \gets 0$
		\WHILE{$\text{Prob}(X, \delta) < 1 - \xi$}
		\FOR{$x_i$ in $X$}
		\IF{$f(w, x_i + \delta) \neq f(w, x_i)$}
		\STATE Solve {\small $\min_{r} \| r \|_2 \text{ s.t. } f(w, x_i + \delta + r) \neq f(w, x_i)$}
		\STATE \qquad by DeepFool
		\STATE Update $\delta \gets \delta + r$,  then project $\delta$ to $\ell_p$ ball
		\ENDIF
		\ENDFOR
		\ENDWHILE
	\end{algorithmic}
\end{algorithm}
Different from \cite{moosavi2017universal}, we consider the following optimization problem for building universal perturbations,
\begin{equation}
    \max_{\delta}  \mcL(w, \delta) =  \frac{1}{N}\sum_{i=1}^{N} l(w, x_i + \delta) \text{ s.t. } \| \delta \|_p \leq \epsilon, \label{eq:univ_prob}
\end{equation}
where $l(w, \cdot)$ represents the loss used for training DNNs.  This simple formulation \eqref{eq:univ_prob} searches for a universal
perturbation that maximizes the training loss, and thus forces images into the wrong class.

The naive formulation \eqref{eq:univ_prob} suffers from a potentially significant drawback; the cross-entropy loss is unbounded from above, and can be arbitrarily large when evaluated on a single image. 
In the worst-case, a perturbation that causes misclassification of just a single image can maximize \eqref{eq:univ_prob} by forcing the average loss to infinity. 
To force the optimizer to find a perturbation that fools many instances, we propose a ``clipped'' version of the cross entropy loss, 
\begin{equation}
\hat l(w, x_i + \delta) = \text{min}\{ l(w, x_i + \delta) , \, \beta\}.
\end{equation}
We cap the loss function at $\beta$ to prevent any single image from dominating the objective in \eqref{eq:univ_prob}, and giving us a better surrogate of misclassification accuracy. 
In \cref{sec:clip_experiment}, we investigate the effect of clipping with different $\beta$.

We directly solve \cref{eq:univ_prob} by a stochastic gradient method described in \cref{alg:univ}. Each iteration begins by using gradient ascent to update the universal perturbation $\delta$ so that it maximizes the loss. Then, $\delta$ is projected onto the $\ell_p$-norm ball to prevent it from growing too large.  We experiment with various optimizers for this ascent step, including Stochastic Gradient Descent (SGD), Momentum SGD (MSGD), Projected Gradient Descent (PGD), and ADAM~\citep{kingma2014adam}. 

\begin{algorithm}
	\caption{
     Stochastic gradient  for universal perturbation
	}
	\label{alg:univ}
	\begin{algorithmic}
		\FOR{epoch $= 1 \ldots N_{ep}$}
		\FOR{minibatch $B\subset X$}
		\STATE Update $\delta$ with gradient variant $\delta \gets \delta + g$
		\STATE Project $\delta$ to $\ell_p$ ball
		\ENDFOR
		\ENDFOR
	\end{algorithmic}
\end{algorithm}

\begin{figure}
    \centering
    \includegraphics[width=0.75\columnwidth]{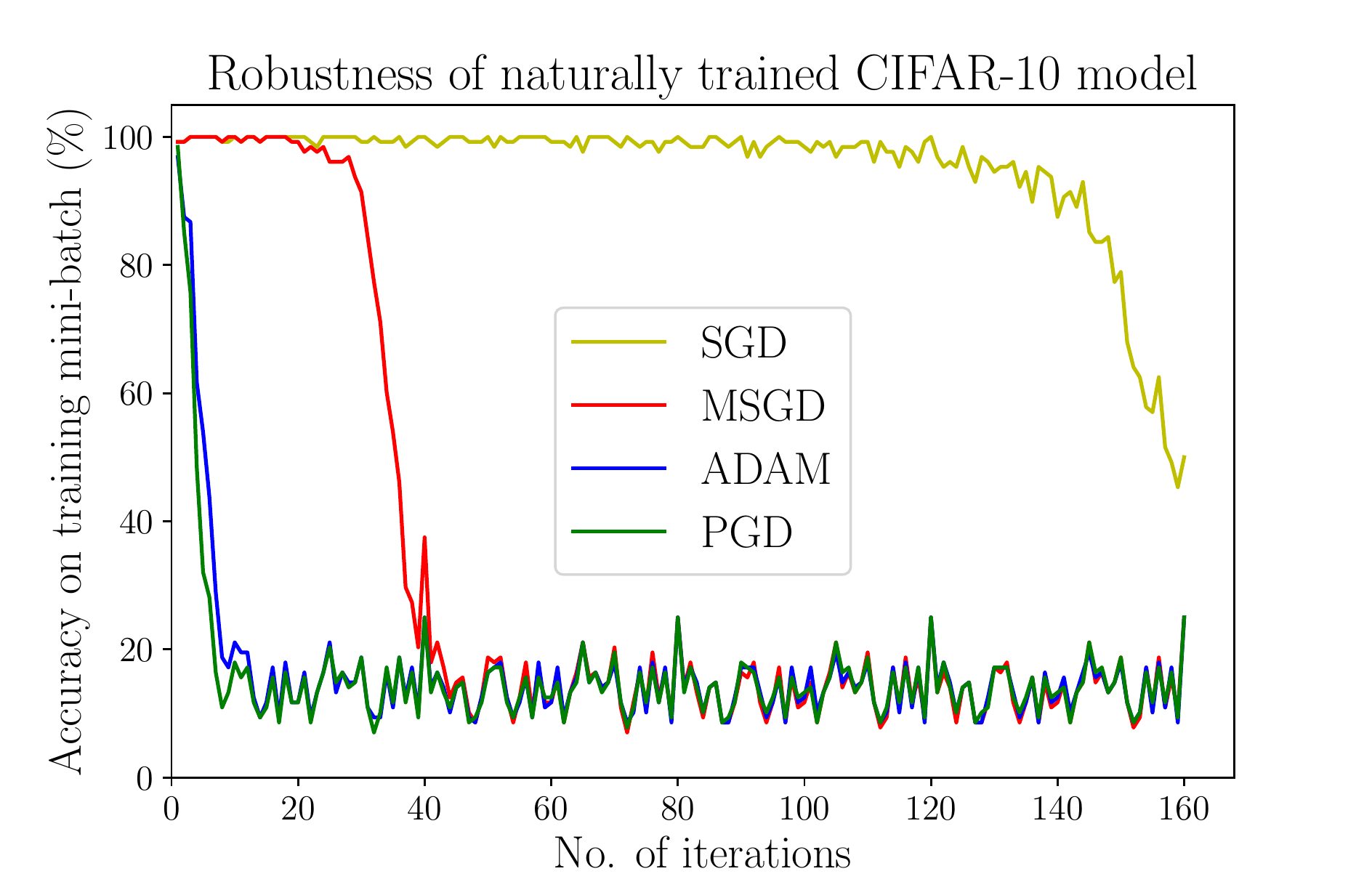}
    \caption{Classification accuracy on adversarial examples of universal perturbations generated by increasing the cross-entropy loss. PGD and ADAM converge faster. We use 5000 training samples from CIFAR-10 for constructing the universal adversarial perturbation for naturally trained WRN model from \cite{madry2017towards}. The batch-size is 128, $\epsilon$=8, and the learning-rate/step-size is 1. 
    }
    \label{fig:cifar10_adversarial_rate}
\end{figure}

\begin{figure}
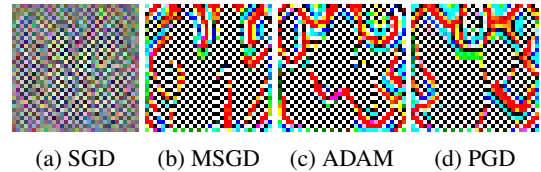

    \centering
    \begin{subfigure}[b]{0.2\columnwidth}
        \centering
        \includegraphics[width=\columnwidth]{sgd_c10_noise.pdf}
        \caption{SGD}
    \end{subfigure}
    \begin{subfigure}[b]{0.2\columnwidth}
        \centering
        \includegraphics[width=\columnwidth]{msgd_c10_noise.pdf}
        \caption{MSGD}
    \end{subfigure}
        \begin{subfigure}[b]{0.2\columnwidth}
        \centering
        \includegraphics[width=\columnwidth]{adam_c10_noise.pdf}
        \caption{ADAM}
    \end{subfigure}
        \begin{subfigure}[b]{0.2\columnwidth}
        \centering
        \includegraphics[width=\columnwidth]{pgd_c10_noise.pdf}
        \caption{PGD}
    \end{subfigure}
    \caption{Visualizations of universal perturbations after 160 iterations of the optimizers depicted in \cref{fig:cifar10_adversarial_rate}.}
    \label{fig:visual_clean_c10} 
\end{figure}

We test this method by attacking a naturally trained WRN 32-10 architecture on the CIFAR-10 dataset. We use $\epsilon=8$ for the $\ell_\infty$ constraint for CIFAR. Stochastic gradient methods that use ``normalized'' gradients (ADAM and PGD) are less sensitive to learning rate and converge faster, as shown in \cref{fig:cifar10_adversarial_rate}. We visualize the generated universal perturbation from different optimizers in \cref{fig:visual_clean_c10}. 
Compared to the noisy perturbation generated
by SGD, normalized gradient methods produced stronger attacks with more well-defined geometric structures and checkerboard patterns.  The final evaluation accuracies (on test-examples) after adding universal perturbations with $\epsilon=8$ were 42.56\% for the SGD perturbation, 13.08\% for MSGD, 13.30\% for ADAM, and 13.79\% for PGD. The clean test accuracy of the WRN is 95.2\%.

Our proposed method of universal attack using a clipped loss function has several advantages.  It is based on a standard stochastic gradient method that comes with convergence guarantees when a decreasing learning rate is used \citep{bottou2018optimization}. 
Also, each iteration is based on a minibatch of samples instead of one instance, which accelerates computation on a GPU.  
Finally, each iteration requires a simple gradient update instead of the complex DeepFool inner loop; we empirically verify fast convergence and good performance of the proposed method (see \cref{sec:exp_opt}).

\section{Universal adversarial training}
\label{sec:univ_adv}
We now consider training robust classifiers that are resistant to universal perturbations.
Similar to \cite{madry2017towards}, we borrow ideas from robust optimization. We use robust optimization to build robust models that can resist universal perturbations. In particular, we consider universal adversarial training, and formulate this problem as a min-max optimization problem,
\begin{equation}
    \min_{w} \max_{\|\delta\|_p \leq \epsilon} \,  \mcL(w,\delta) =  \frac{1}{N}\sum_{i=1}^{N} l(w, x_i + \delta)
\label{eq:adv_univ}
\end{equation}
where $w$ represents the neural network weights, $X=\{x_i, i=1,\ldots,N\}$ represents training samples, $\delta$ represents universal perturbation noise, and $l(\cdot)$ is the loss function. Here, unlike conventional adversarial training, our $\delta$ is a universal perturbation (or, more accurately, mini-batch universal).
Previously, solving this optimization problem directly was deemed computationally infeasible due to the large cost associated with generating a universal perturbation \citep{perolat2018playing}, but we show that \cref{eq:adv_univ} is efficiently solvable by alternating stochastic gradient methods shown in \cref{alg:adv_univ}.  
We show that unlike \cite{madry2017towards}, updating the universal perturbation only using a simple step is enough for building universally hardened networks. Each iteration alternatively updates the neural network weights $w$ using gradient descent, and then updates the universal perturbation $\delta$ using ascent. 

\begin{algorithm}
	\caption{
		Alternating stochastic gradient method for adversarial training against universal perturbation
	}
	\label{alg:adv_univ}
	\begin{algorithmic}
		\REQUIRE Training samples $X$, perturbation bound $\epsilon$, learning rate $\tau$, momentum $\mu$
		\FOR{epoch $= 1 \ldots N_{ep}$}
		\FOR{minibatch $B\subset X$}
		\STATE Update $w$ with momentum stochastic gradient
		\STATE \qquad  $g_w \gets  \mu g_w  - \bbE_{x \in B} [\nabla_w \, l(w, x + \delta)]$
		\STATE \qquad  $w \gets w + \tau g_w$ 
		\STATE Update $\delta$ with  stochastic gradient ascent 
		\STATE \qquad $\delta \gets \delta + \epsilon \, \text{sign}(\bbE_{x \in B} [\nabla_{\delta} \, l(w, x + \delta)])$ 
		\STATE Project $\delta$ to $\ell_p$ ball
		\ENDFOR
		\ENDFOR
	\end{algorithmic}
\end{algorithm}

We compare our formulation \eqref{eq:adv_univ} and \cref{alg:adv_univ} with PGD-based adversarial training, which trains a robust model by optimizing the following min-max problem,
\begin{equation}
    \min_{w} \max_{Z} \,   \frac{1}{N}\sum_{i=1}^{N} l(w, z_i)  \quad \text{ s.t. }  \| Z - X \|_p \leq \epsilon. 
 \label{eq:madry}
\end{equation}
The standard formulation \eqref{eq:madry} searches for per-instance perturbed images $Z$, while our formulation in \eqref{eq:adv_univ} maximizes using a universal perturbation $\delta$. 
\cite{madry2017towards} solve \eqref{eq:madry} by a stochastic method. In each iteration, an adversarial example $z_i$ is generated for an input instance by the PGD iterative method, and the DNN parameter $w$ is updated once. 
Our formulation (\cref{alg:adv_univ}) only maintains one single perturbation that is used and refined across all iterations.  For this reason, we need only update $w$ and $\delta$ once per step (there is no expensive inner loop), and these updates accumulate for both $w$ and $\delta$ through training.

We consider different rules for updating $\delta$ during universal adversarial training,
\begin{align}
    \text{FGSM  } & \delta \gets \delta + \epsilon  \cdot \text{sign}(\bbE_{x \in B} [\nabla_{\delta} l(w, x + \delta)]), \\
    \text{SGD  } & \delta \gets \delta + \tau_{\delta}\cdot \bbE_{x \in B} [\nabla_{\delta} l(w, x + \delta)],
\end{align}
and ADAM.
We found that the FGSM update rule was most effective when combined with the SGD optimizer for updating DNN weights $w$.

We use fairly standard training parameters in our experiments. In our CIFAR experiments, we use $\epsilon=8$, batch-size of 128, and we train for 80,000 steps. For the optimizer, we use Momentum SGD with an initial learning rate of 0.1 which drops to 0.01 at iteration 40,000 and drops further down to 0.001 at iteration 60,000. 
One way to assess the update rule is to plot the model accuracy before and after the ascent step (i.e., the perturbation update).  It is well-known that adversarial training is more effective when stronger attacks are used. In the extreme case of a do-nothing adversary, the adversarial training method degenerates to natural training. 
As illustrated in the supplementary, we see a gap between the accuracy curves plotted before and after gradient ascent.
We find that the FGSM update rule leads to a larger gap, indicating a stronger adversary. 
Correspondingly, we find that the FGSM update rule yields networks that are more robust to attacks as compared to SGD update.

\subsection{Attacking hardened models}
We evaluate the robustness of different models by applying \cref{alg:univ} to find universal perturbations. We attack universally adversarial trained models (produced by \cref{eq:adv_univ}) using the FGSM universal update rule (uFGSM), or the SGD universal update rule (uSGD).  We also consider robust models from per-instance adversarial training (\cref{eq:madry}) with adversarial steps of the FGSM and PGD type. 

\begin{figure}[t]
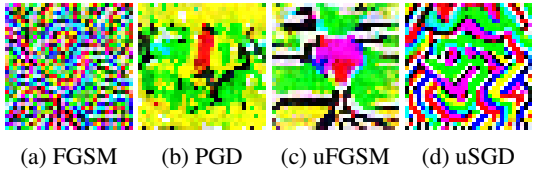

    \centering
    \begin{subfigure}[b]{0.2\columnwidth}
        \centering
        \includegraphics[width=\columnwidth]{FGSM_madry_c10_pgd.pdf}
        \caption{FGSM}
    \end{subfigure}
    \begin{subfigure}[b]{0.2\columnwidth}
        \centering
        \includegraphics[width=\columnwidth]{madry_c10_pgd.pdf}
        \caption{PGD}
    \end{subfigure}
        \begin{subfigure}[b]{0.2\columnwidth}
        \centering
        \includegraphics[width=\columnwidth]{uAdvTFGSMMax_c10_pgd.pdf}
        \caption{uFGSM}
    \end{subfigure}
        \begin{subfigure}[b]{0.2\columnwidth}
        \centering
        \includegraphics[width=\columnwidth]{uAdvTsgdMax_c10_pgd.pdf}
        \caption{uSGD}
    \end{subfigure}
    \caption{Universal perturbations made (with 400 iterations) 
    for 4 different CIFAR-10 robust models: adversarially trained with FGSM or PGD, and universally adversarially trained with FGSM (uFGSM) or SGD (uSGD). 
    }
    \label{fig:visual_advTrained_c10}
\end{figure}

Robust models adversarially trained with weaker attackers such as uSGD and (per-instance) FGSM are relatively vulnerable to universal perturbations, while robust models from (per-instance) PGD and uFGSM can resist universal perturbations. 
We plot the universal perturbations generated using \cref{alg:adv_univ} in \cref{fig:visual_advTrained_c10}. 
When we visually compare the universal perturbations of robust models (\cref{fig:visual_advTrained_c10}) and those of a naturally trained model (\cref{fig:visual_clean_c10}), we can see a drastic change in structure. 
Similarly, even among hardened models, universal perturbations generated for weaker robust models have more geometric textures, as shown in \cref{fig:visual_advTrained_c10} 
(a,d).

While an $\epsilon$-bounded per-instance robust model is also robust against $\epsilon$-bounded universal attacks since the universal attack is a more constrained version of the per-instance attack, training robust per-instance models is only possible for small datasets like CIFAR and for small $\epsilon$. For larger datasets like ImageNet, we cannot train per-instance robust models with such large $\epsilon$'s common for universal attacks. However, we include the per-instance adversarially trained model as a candidate universally robust model for CIFAR-10 in our comparisons to allow comparisons in settings where it is possible to train per-instance robust models.   

We apply the strongest attack to validation images of the natural model and various universal adversarially trained models using different update steps. The results are summarized in \cref{tab:c10_uadv}. Our models become robust against universal perturbations and have higher accuracies on natural validation examples compared to per-instance adversarially trained models. Their robustness is even more if attacked with iDeepFool (93.29\%). Compared to the per-instance FGSM trained model which has the same computational cost as ours, our universal model is more robust. Note that the PGD trained model is trained on a 7-step per-instance adversary and requires about $4\times$ more computation than ours. 

\begin{table}
\centering
\caption{Validation accuracy of hardened WideResnet models trained on CIFAR-10. Note that Madry's PGD training is significantly slower than the other training methods. 
} 
\setlength\tabcolsep{3pt}
\begin{tabular}{|c|c|c|c|}
\hline
\multicolumn{2}{|c|}{} & \multicolumn{2}{c|}{Validation Accuracy on} \\
\cline{3-4}
\multicolumn{2}{|c|}{} & UnivPert & Natural  \\
\hline
\multirow{4}{*}{\tabincell{c}{(Robust) \\ models \\ trained \\ with}} & Natural & 9.2\% & 95.2\% \\
\cline{2-4}
& FGSM & 51.0\% & 91.42\%\\
\cline{2-4}
& PGD & 86.1\% & 87.25\%\\
\cline{2-4}
& uADAM (ours) & 91.6\% & \textbf{94.28\%}\\
\cline{2-4}
& uFGSM (ours) & \textbf{91.8\%} & 93.50\%\\
\hline
\end{tabular}%
\label{tab:c10_uadv}%
\end{table}%

\subsection{Low-cost universal adversarial training}
As shown in \cref{tab:c10_uadv}, our proposed algorithm \ref{alg:adv_univ} was able to harden the CIFAR-10 classification network. This comes at the cost of doubling the training time. Adversarial training in general should have some cost since it requires the generation or update of the adversarial perturbation of the mini-batch {\em before} each minimization step on the network's parameters. However, since universal perturbations are approximately image-agnostic, results should be fairly invariant to the order of updates.  For this reason, we propose to compute the image gradient needed for the perturbation update during the same backward pass used to compute the parameter gradients.  This results in a simultaneous update for network weights and the universal perturbation in \cref{alg:free_adv_univ}, which backprops only once per iteration and produces approximately universally robust models at almost no cost in comparison to natural training. The ``low-cost universal adversarially trained'' model of CIFAR-10 is 86.1\% robust against universal perturbations and has 93.5\% accuracy on the clean validation examples. When compared to the original version in \cref{tab:c10_uadv}, the robustness has only slightly decreased. However, the training time is cut by half. This is a huge improvement in efficiency, in particular for large datasets like ImageNet with long training times.

\begin{algorithm}
	\caption{
		Simultaneous stochastic gradient method for adversarial training against universal perturbation
	}
	\label{alg:free_adv_univ}
	\begin{algorithmic}
		\REQUIRE Training samples $X$, perturbation bound $\epsilon$, learning rate $\tau$, momentum $\mu$
		\STATE Initialize $w, \delta$
		\FOR{epoch $= 1 \ldots N_{ep}$}
		\FOR{minibatch $B\subset X$}
		\STATE Compute gradient of loss with respect to $w$ and $\delta$
		\STATE \qquad $d_w \gets \bbE_{x \in B} [\nabla_w \, l(w, x + \delta)]$
		\STATE \qquad $d_\delta \gets \bbE_{x \in B} [\nabla_\delta \, l(w, x + \delta)]$
		\STATE Update $w$ with momentum stochastic gradient
		\STATE \qquad  $g_w \gets  \mu g_w  - d_w$
		\STATE \qquad  $w \gets w + \tau g_w$ 
		\STATE Update $\delta$ with  stochastic gradient ascent 
		\STATE \qquad $\delta \gets \delta + \epsilon \text{sign}(d_\delta)$ 
		\STATE Project $\delta$ to $\ell_p$ ball
		\ENDFOR
		\ENDFOR
	\end{algorithmic}
\end{algorithm}

\section{Universal perturbations for ImageNet}
\label{sec:exp_opt}
To validate the performance of our proposed optimization on different architectures and more complex datasets, we apply \cref{alg:univ} to various popular architectures designed for classification on the ImageNet dataset \citep{russakovsky2015imagenet}. 
We compare our method of universal perturbation generation with the current state-of-the-art method, iterative DeepFool (iDeepFool for short -- alg.~\ref{alg:cvpr}). We use the authors' code to run the iDeepFool attack on these classification networks.
We execute both our method and iDeepFool on the exact same 5000 training data points and terminate both methods after 10 epochs. 
We use $\epsilon=10$ for $\ell_\infty$ constraint following \cite{moosavi2017universal}, use a step-size of 1.0 for our method, and use suggested parameters for iDeepFool. Similar conclusions could be drawn when we use $\ell_2$ bounded attacks.  
We independently execute iDeepFool since we are interested in the accuracy of the classifier on attacked images -- a metric not reported in \cite{moosavi2017universal} \footnote{They report ``fooling ratio'' which is the ratio of examples who's label prediction changes after applying the universal perturbation. This has become an uncommon metric since the fooling ratio can increase if the universal perturbation causes an example that was originally miss-classified to become correctly classified.}.
\paragraph{Benefits of the proposed method}
We compare the performance of our stochastic gradient method for  \cref{eq:univ_prob} and the iDeepFool method for \cref{eq:cvpr}.
We generate universal perturbations for Inception \citep{szegedy2016rethinking} and VGG \citep{simonyan2014very} networks trained on ImageNet, and report the top-1 accuracy in \cref{tab:opt}. 
Universal perturbations generated by both iDeepFool and our method fool networks and degrade the classification accuracy.  Universal perturbations generated for the training samples generalize well and cause the accuracy of validation samples to drop.  However, when given a fixed computation budget such as number of passes on the training data, our method outperforms iDeepFool by a large margin. Our stochastic gradient method generates the universal perturbations at a much faster pace than iDeepFool. About 20$\times$ faster on InceptionV1 and 6$\times$ on VGG16 (13$\times$ on average). 

After verifying the effectiveness and efficiency of our proposed stochastic gradient method\footnote{Unless otherwise specified, we use the sign-of-gradient PGD for our stochastic gradient optimizer in \cref{alg:univ}.}, we use our \cref{alg:univ} to generate universal perturbations for ResNet-V1 152 \citep{he2016deep}  and Inception-V3. Our attacks degrade the validation accuracy of ResNet-V1 152 and Inception-V3 from 76.8\% and 78\% to 16.4\% and 20.1\%, respectively. 
The final universal perturbations used for the results presented  are illustrated in the supplementary.

\begin{table}[h]
    \centering
    \caption{Top-1 accuracy on ImageNet for natural images, and adversarial images with universal perturbation. 
    } 
    \begin{tabular}{|c|c|c|c|}
    \hline
    \multicolumn{2}{|c|}{} & InceptionV1 & VGG16 \\ 
    \hline
    \multirow{2}{*}{Natural} & Train & 76.9\% & 81.4 \% \\
    & Val & 69.7\% & 70.9\%\\
    \hline
    \multirow{2}{*}{iDeepFool} & Train & 43.5\% & 39.5\% \\
    & Val & 40.7\% & 36.0\% \\
    \hline
    \multirow{2}{*}{Ours} & Train & \textbf{17.2\%} & \textbf{23.1\%} \\
    & Val & \textbf{19.8\%} & \textbf{22.5\%}\\
    \hline
    \multicolumn{2}{|c|}{iDeepFool time (s)} & 9856 & 6076 \\
    \hline
    \multicolumn{2}{|c|}{our time (s)} & \textbf{482} & \textbf{953} \\
    \hline
    \end{tabular}
    \label{tab:opt}
    \vspace{-2mm}
\end{table}

\begin{figure}[h]
    \centering
    \includegraphics[width=0.7\columnwidth]{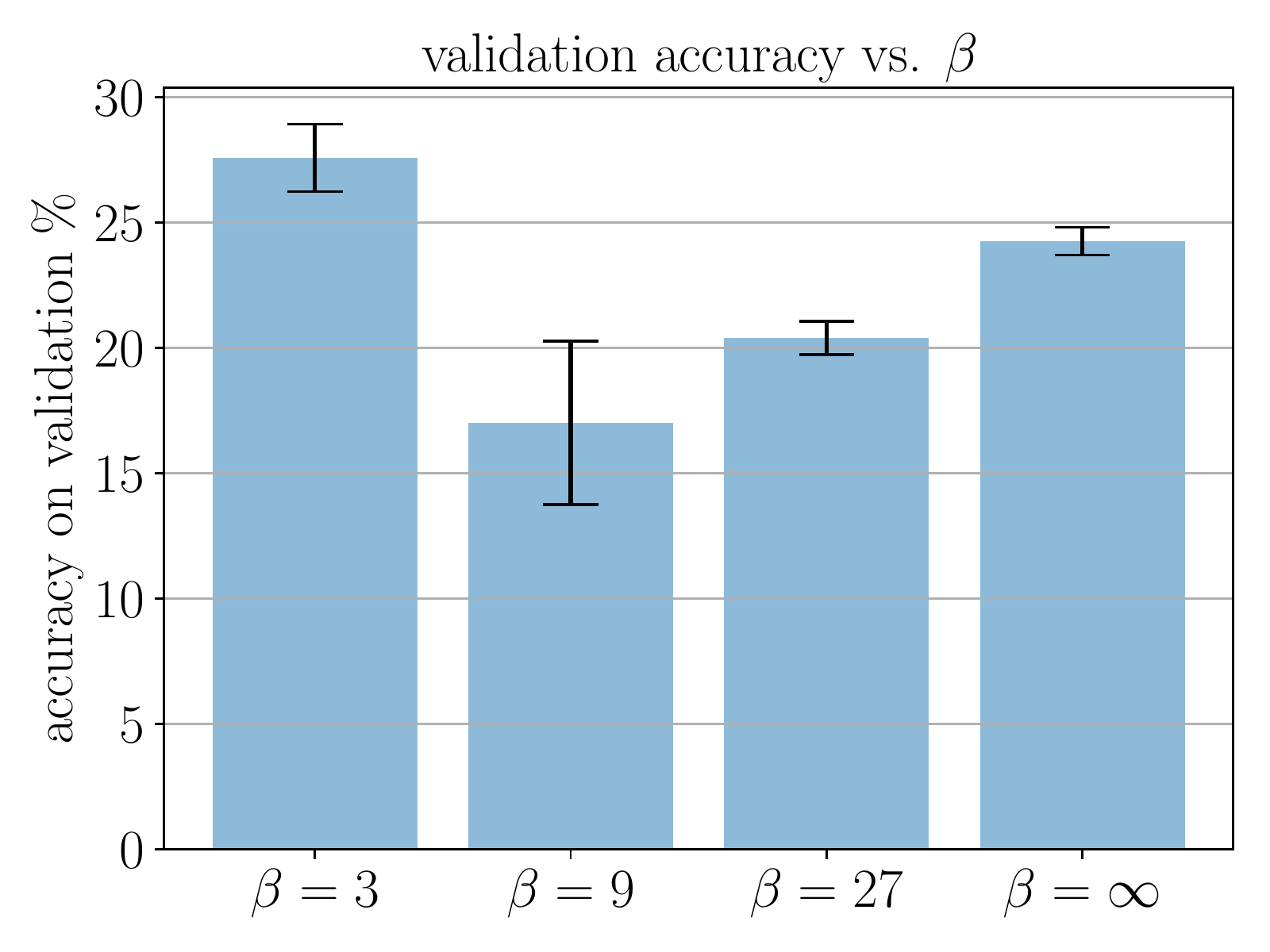}
    \caption{Attack performance varies with clipping parameter $\beta$ in \cref{eq:univ_prob}. 
    Attacking Inception-V3 is more successful with clipping ($\beta=9$) than without clipping ($\beta=\infty$).
    }
    \label{fig:beta}
        \vspace{-2mm}
\end{figure}

\paragraph{The effect of clipping}\label{sec:clip_experiment}
Here, we analyze the effect of the ``clipping'' loss parameter $\beta$ in \cref{eq:univ_prob}. For this purpose, similar to our other ablations, we generate universal perturbations by solving \cref{eq:univ_prob} using PGD for Inception-V3 on ImageNet. 
We run each experiment with 5 random subsets of training data. The accuracy reported is the classification accuracy on the entire validation set of ImageNet after adding the universal perturbation. The results are summarized in \cref{fig:beta}. The results showcase the value of our proposed loss function for finding universal adversarial perturbations.

\begin{figure}[h]
    \centering
    \includegraphics[width=0.7\columnwidth]{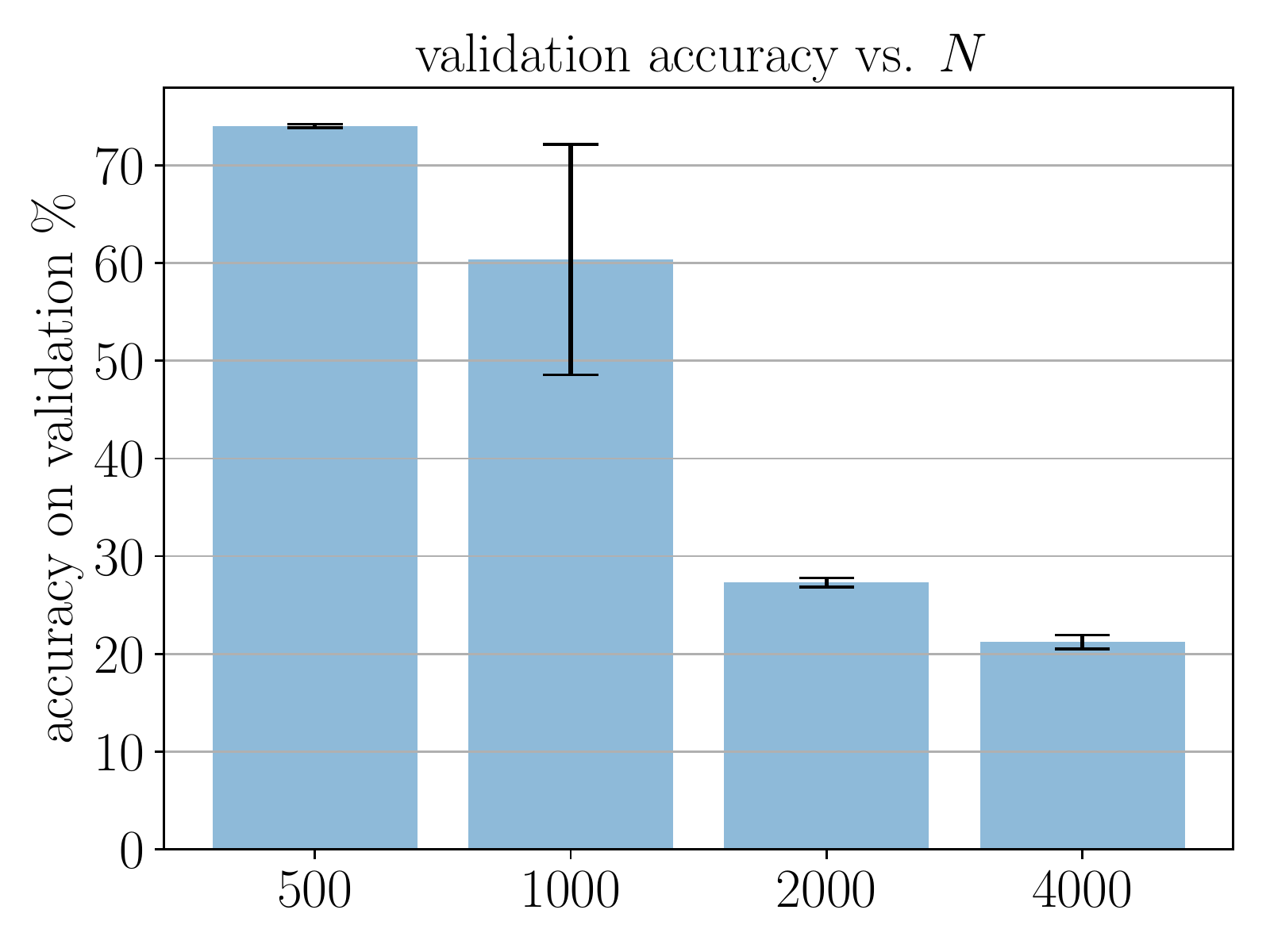}
    \caption{Attack success significantly improves when the number of training points is larger than number of classes. For reference, Inception-V3's top-1 accuracy is 78\%. $N_{ep}$ in \cref{alg:univ} was set to 100 epochs for 500 data samples, 40 for 1000 and 2000 samples, and 10 for more.}
    \label{fig:numTrain}
        \vspace{-6mm}
\end{figure}

\begin{figure*}[t]
    \centering
    \begin{subfigure}{0.63\columnwidth}
        \includegraphics[width=\columnwidth, trim=0cm 0cm 0cm 1.3cm, clip]{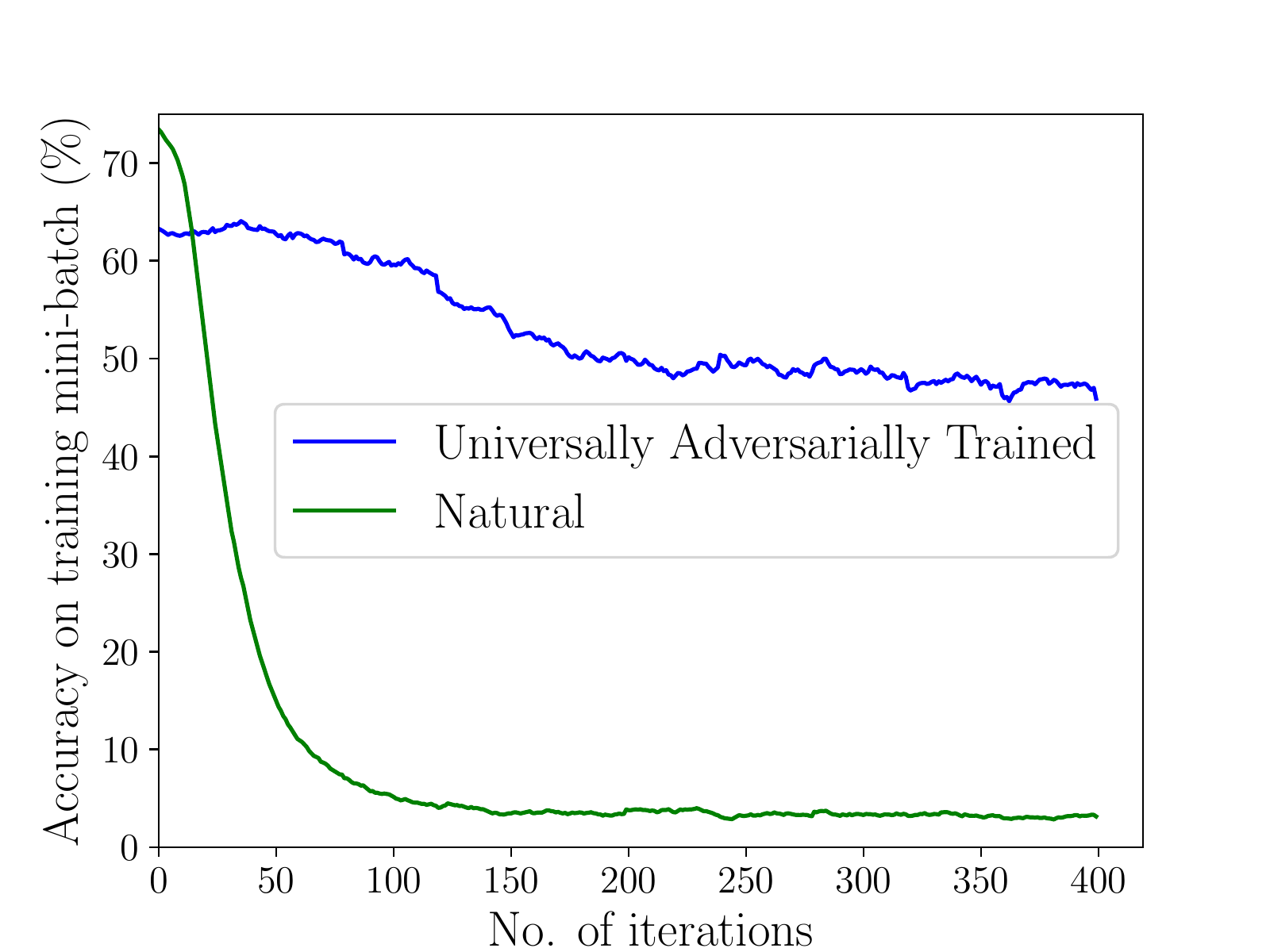}
        \caption{Generating universal perturbations}
    \end{subfigure}
    \begin{subfigure}{0.43\columnwidth}
        \includegraphics[width=\columnwidth]{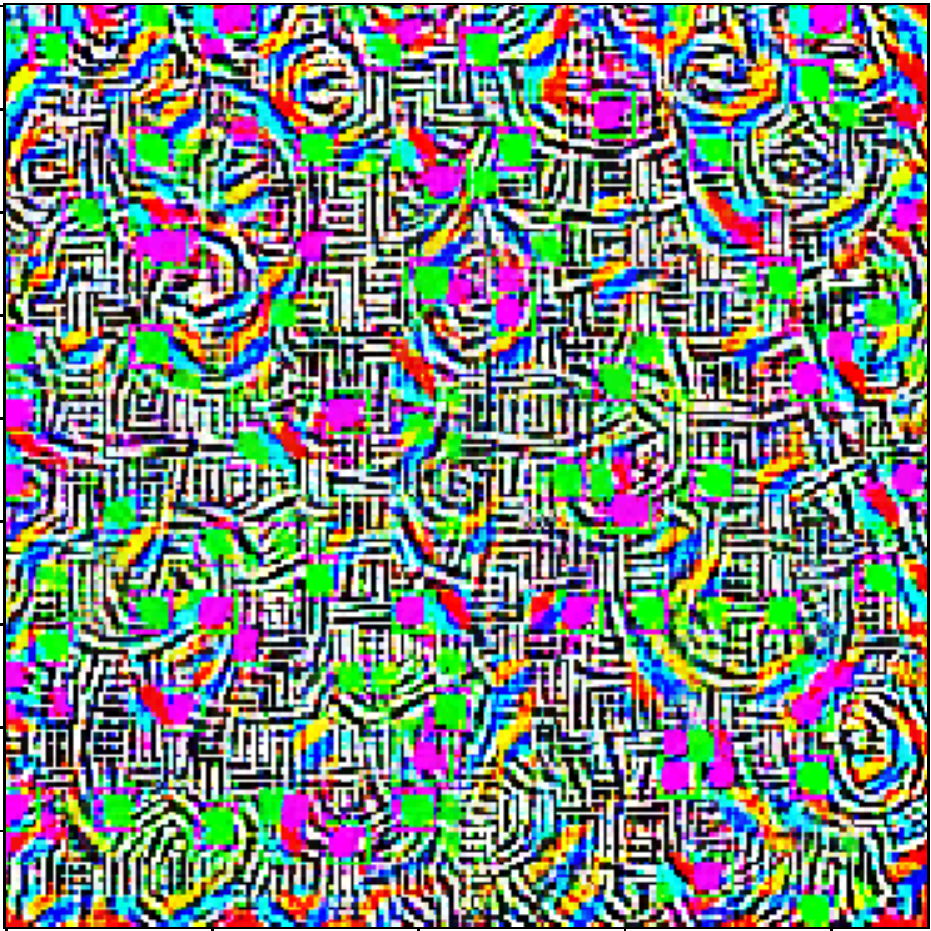}
        \caption{\footnotesize Natural AlexNet}
    \end{subfigure}
    \begin{subfigure}{0.43\columnwidth}
        \includegraphics[width=\columnwidth]{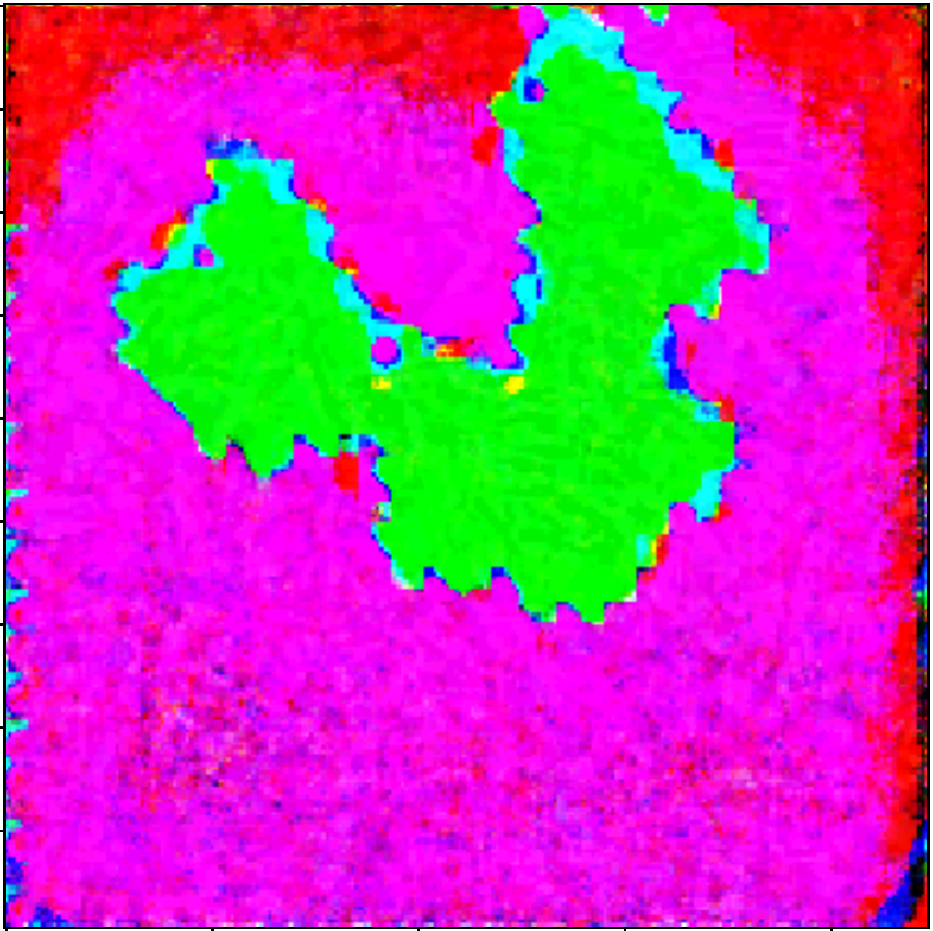}
        \caption{\footnotesize  Universal Training}
    \end{subfigure}
        \begin{subfigure}{0.43\columnwidth}
        \includegraphics[width=\columnwidth]{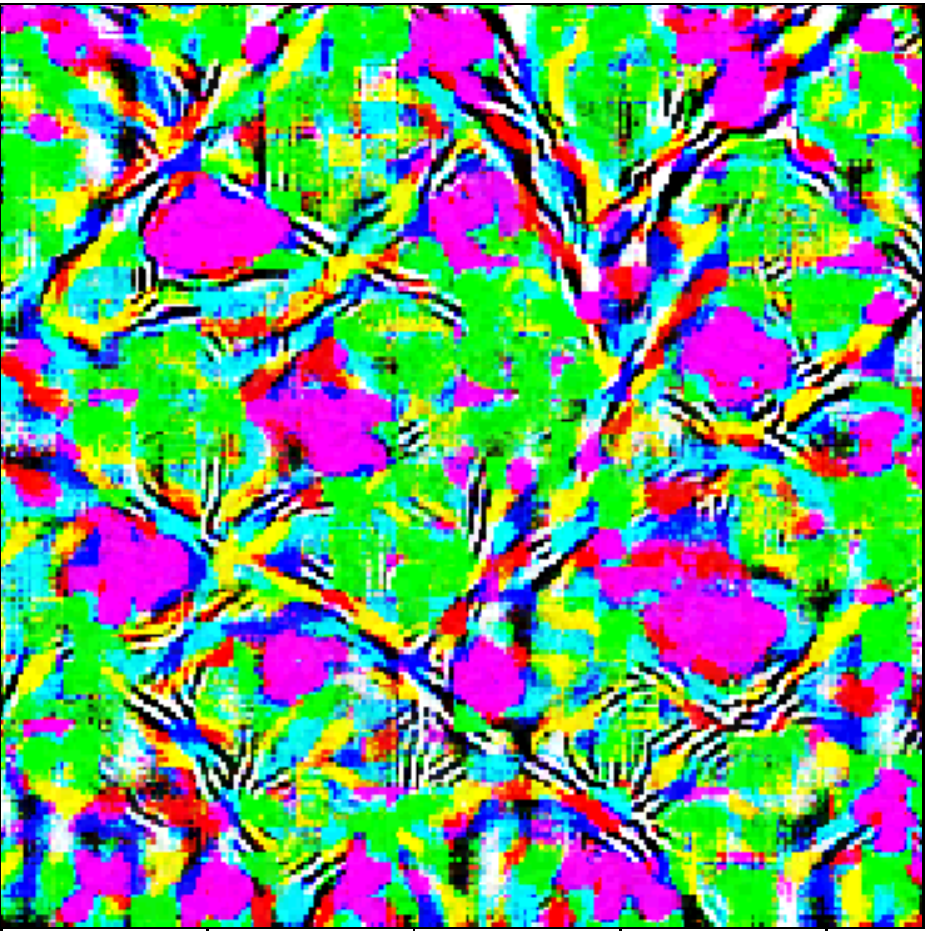}
        \caption{\footnotesize  Low-cost Universal Tr.}
    \end{subfigure}
    \caption{Universal perturbations fool naturally trained AlexNet on ImageNet, but fail to fool our robust AlexNets.
    The universal perturbations generated for the universal adversarial trained AlexNets have little geometric structure compared to that of naturally trained nets.  (b) Universal perturbation of natural model. The accuracy of the validation images + noise is only \textbf{3.9}\% (c)  Perturbation for our universally trained model using \cref{alg:adv_univ}. The accuracy of the validation images + noise is \textbf{42.0\%}. (d)  Perturbation for the model trained with low-cost universal training variant (\cref{alg:free_adv_univ}). The accuracy of the validation images + noise is \textbf{28.3}\%. While the universal noise for the low-cost variant of universal adversarial training has some structure compared to the original, it is less structured than an attack on the natural model (b). Curves smoothed for better visualization. }
    \label{fig:AlexNetNoise}
\end{figure*}

\paragraph{How much training data does the attack need?}
As in \cite{moosavi2017universal}, we analyze how the number of training points ($|X|$) affects the strength of universal perturbations in \cref{fig:numTrain}. We build $\delta$ using varying amounts of training data. For each experiment, we report the accuracy on the entire validation set after we add the perturbation $\delta$.  

\section{Universal adversarial training ImageNet}

In this section, we analyze our robust models that are universal adversarially trained by solving the min-max problem (\cref{sec:univ_adv}) using \cref{alg:adv_univ}. 
We use $\epsilon=10$ for ImageNet. Note that unlike CIFAR where we were able to train a per-instance PGD-7 robust model with $\epsilon=8$, for ImageNet, there exists no model which can resist per-instance non-targeted perturbations with such large $\epsilon$. For ImageNet, we again use fairly standard training parameters (90 epochs, batch-size 256).

Since our universal adversarial training algorithm (\cref{alg:adv_univ}) is cheap, it scales to large datasets such as ImageNet. We first train an AlexNet model on ImageNet. We use the natural training hyper-parameters for universal adversarially training our AlexNet model. Also, we separately use our ``low-cost universal training'' algorithm to train a robust AlexNet with no overhead cost. We then attack the natural, universally trained, and  no-cost universally trained versions of AlexNet using universal attacks.

As seen in \cref{fig:AlexNetNoise} (a), the AlexNet trained using our universal adversarial training algorithm (\cref{alg:adv_univ}) is robust against universal attacks generated using both \cref{alg:cvpr} and \cref{alg:univ}.
The naturally trained AlexNet is susceptible to universal attacks. The final attacks generated for the robust and natural models are presented in \cref{fig:AlexNetNoise} (b,c,d). The universal perturbation generated for the robust AlexNet model has little structure compared to the universal perturbation built for the naturally trained AlexNet. This is similar to the trend we observed in \cref{fig:visual_clean_c10} and \cref{fig:visual_advTrained_c10} for the WRN models trained on CIFAR-10. 

The accuracy of the universal perturbations on the validation examples are summarized in \cref{tab:UAT_IMGNET}. Similar to CIFAR-10, the low-cost version of universal adversarial training is robust but not as robust as the main method.
\begin{table}[h]
    \centering
    \caption{Accuracy on ImageNet for nat and robust models.
    } 
    \begin{tabular}{|c||c|c|c|c|}
    \hline
     \multirow{3}{*}{\textbf{Training}} & \multicolumn{4}{c|}{ Evaluated Against} \\ \cline{2-5} & \multicolumn{2}{c|}{Natural Images} & \multicolumn{2}{c|}{Universal Attack} \\ \cline{2-5} & Top-1 & Top-5 & Top-1 & Top-5 \\
    \hline\hline
    Natural & 56.4\% & 79.0 \% & 3.9 \% & 9.4 \%\\
    \hline
    Universal& 49.5\% & 72.7\% & 42.0\% & 65.8 \% \\
    \hline
    Low-cost U.& 48.4\% & 72.4\% & 28.3\% & 48.3 \% \\
    \hline
    \end{tabular}
    \label{tab:UAT_IMGNET}
\end{table}
We also train a universally robust ResNet-101 ImageNet model. While a naturally trained ResNet-101 achieves only 7.23\% accuracy on universal perturbations, our ResNet-101 achieves 74.43\% top1 and 92.00\% top5 accuracies.

\section{Conclusion}
We proposed using stochastic gradient methods and a ``clipped'' loss function as an effective universal attack that generates universal perturbations much faster than previous methods.
To defend against universal perturbations, we proposed to train robust models by optimizing a min-max problem using alternating or simultaneous stochastic gradient methods. We show that this is possible using certain universal noise update rules that use ``normalized'' gradients.
The simultaneous stochastic gradient method comes at almost no extra cost compared to natural training and has almost no additional cost compared to conventional training. 
\paragraph{Acknowledgements} Goldstein and his students were supported by the DARPA GARD, DARPA QED for RML, and DAPRA YFA programs. Further support came from the AFOSR MURI program. 
Davis and his students were supported by the DARPA MediFor program under cooperative agreement FA87501620191, “Physical and Semantic Integrity Measures for Media Forensics”. 
Dickerson was supported by NSF CAREER Award IIS-1846237 and DARPA SI3-CMD Award S4761.
The views and conclusions contained herein are those of the authors and should not be interpreted as necessarily representing the official policies or endorsements, either expressed or implied, of the ODNI, IARPA, or the U.S. Government. The U.S. Government is authorized to reproduce and distribute reprints for Governmental purposes notwithstanding any copyright annotation thereon.
{\small
\bibliographystyle{aaai}
\bibliography{adv}
}
\clearpage
\appendix
\newpage
\FloatBarrier
\section{Training curves for UAT}
\begin{figure}[t]
    \centering
    \begin{subfigure}[t]{\columnwidth}
        \includegraphics[width=\columnwidth]{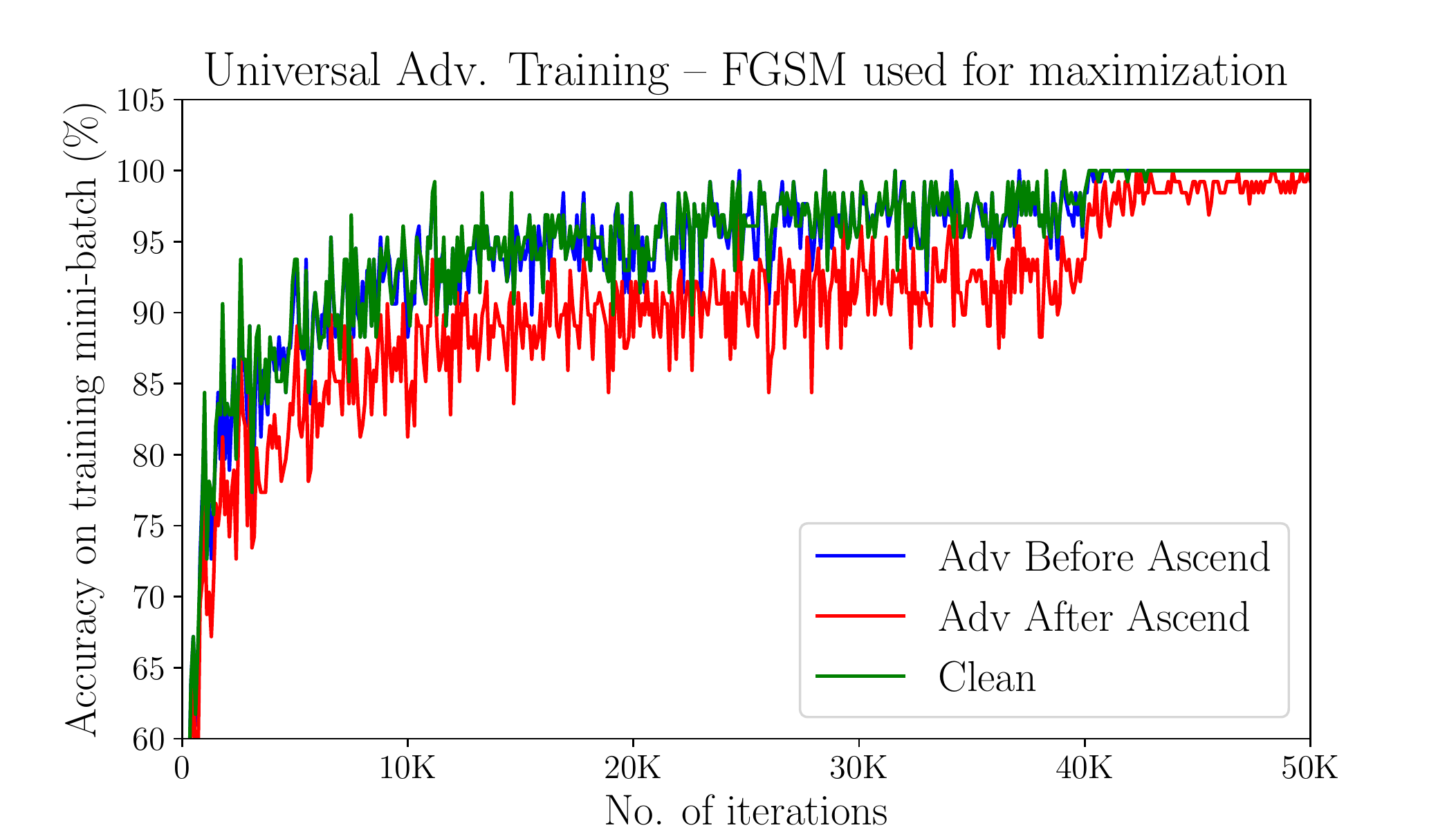}
    \end{subfigure}
    \begin{subfigure}[t]{\columnwidth}
    \includegraphics[width=\columnwidth]{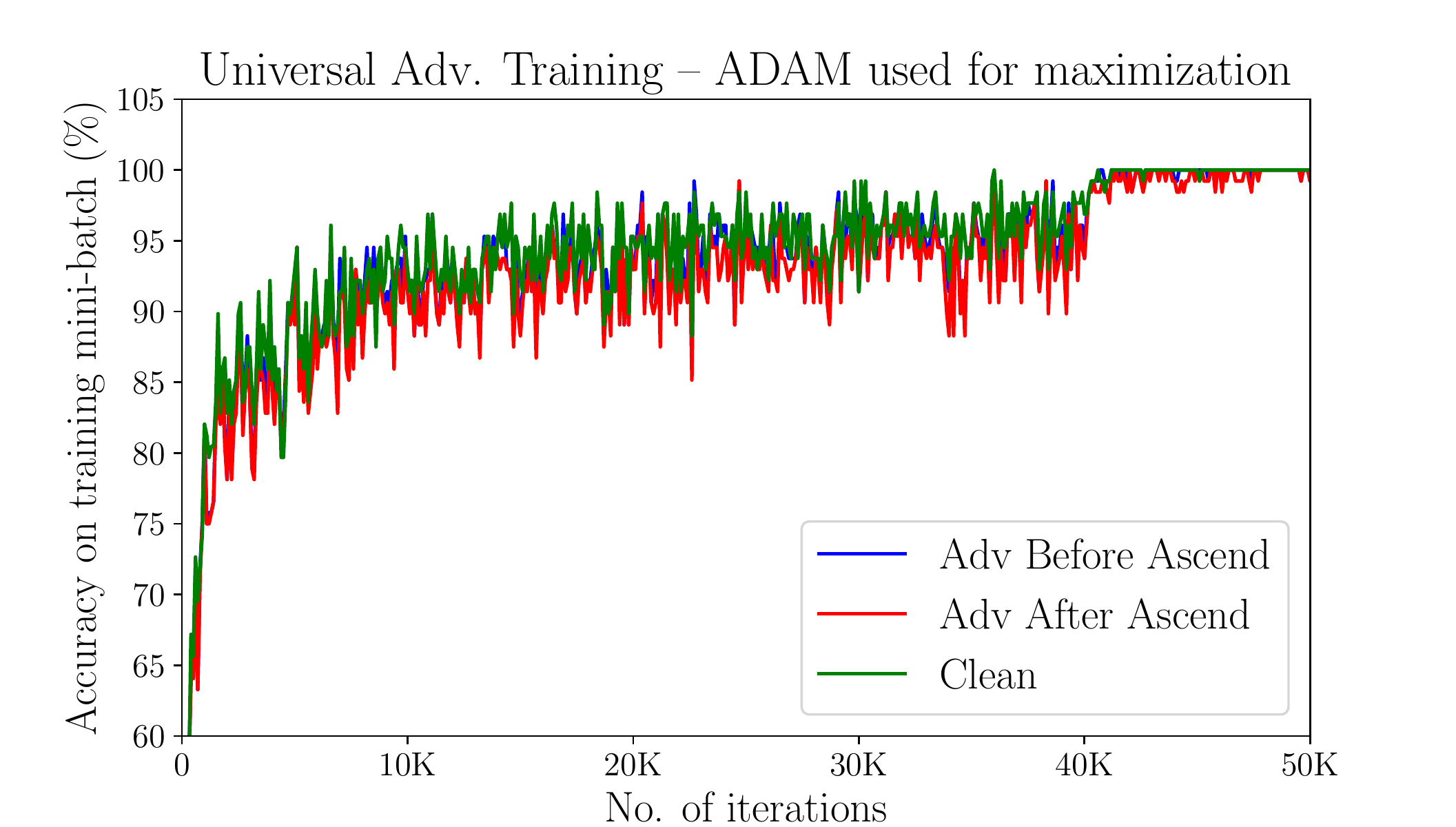}
    \end{subfigure}
    \caption{Classification accuracy for (adversarial) training of (robust) models with (top) FGSM update  and (bottom) ADAM update. We show the accuracy before and after the gradient ascent for $\delta$ in \cref{alg:adv_univ}. We omitted the figure for SGD update because the gap between the two curves for SGD is invisible. 
    }
    \label{fig:training_FGSM}
\end{figure}

In \cref{fig:training_FGSM}, we present training curves for the CIFAR-10 universal adversarial training process on the WideResnet 32-10 architecture. 

\begin{figure}[!h]
    \centering
    \includegraphics[width=\columnwidth]{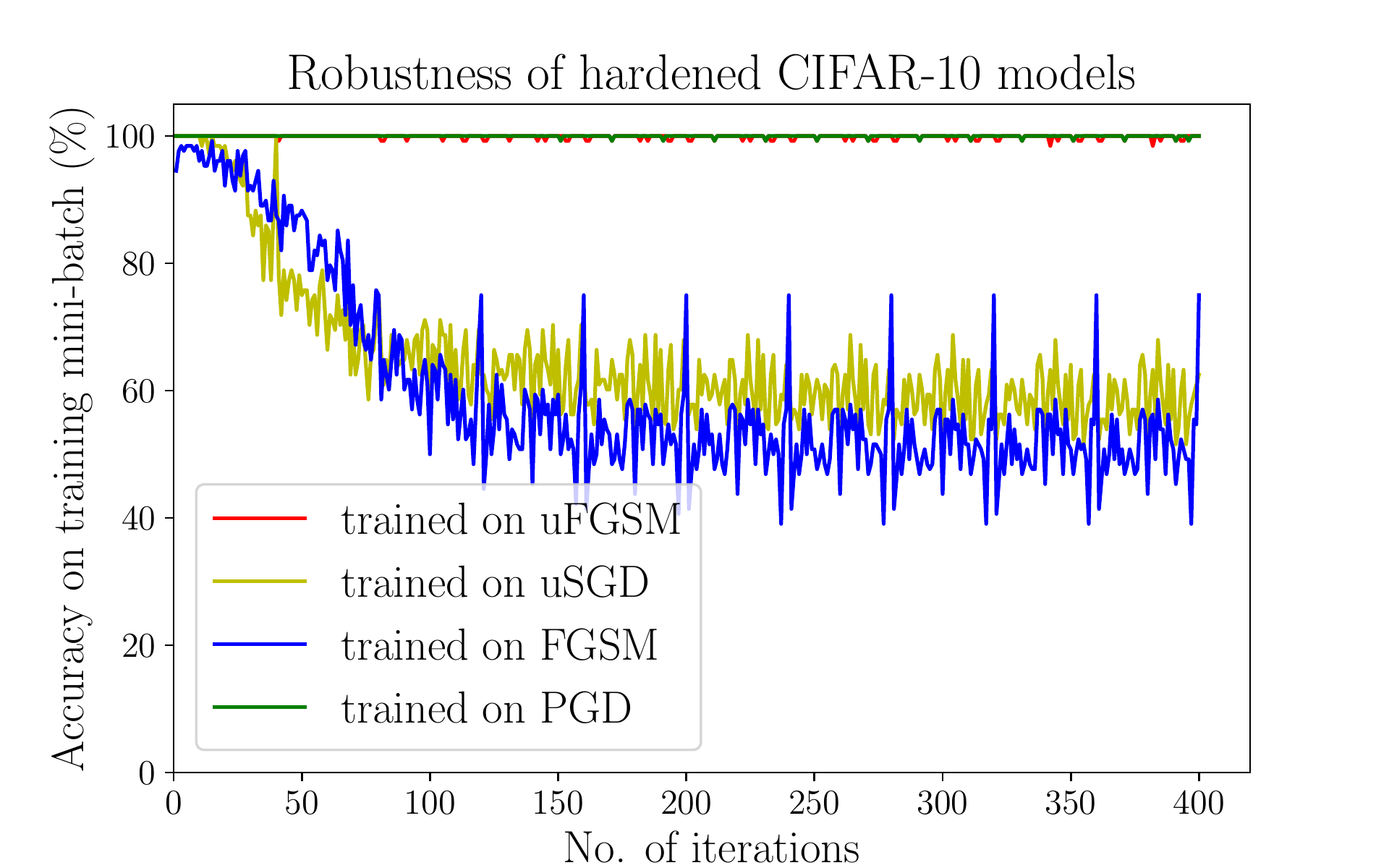}
    \caption{Classification accuracy on training data when the universal perturbations are updated with the ADAM optimizer. We use 5000 training samples from CIFAR-10 for constructing the universal adversarial perturbation for an adversarially trained WideResnet model from \cite{madry2017towards}. The batch-size is 128, $\epsilon$=8, and the lr/step-size is 1.}
    \label{fig:cifar10_robust_accuracies}
\end{figure}

As seen in \cref{fig:training_FGSM}, the gap before and after ascent is largest when the FGSM update rule for universal perturbations is used. Also, in \cref{fig:cifar10_robust_accuracies} we can see that the universal adversarially trained model that uses the FGSM update rule (uFGSM) for its maximization step yeilds the most robust model. 

\section{Universal perturbations using different optimizers}
\begin{figure}[tbhp]
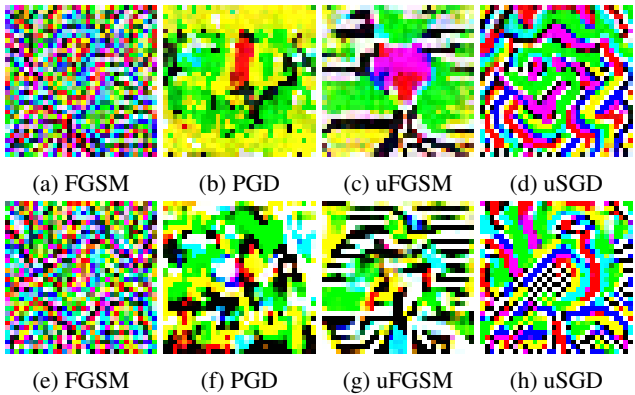

    \centering
    \begin{subfigure}[b]{0.24\columnwidth}
        \includegraphics[width=\columnwidth]{FGSM_madry_c10_pgd.pdf}
        \caption{FGSM}
    \end{subfigure}
    \begin{subfigure}[b]{0.24\columnwidth}
        \includegraphics[width=\columnwidth]{madry_c10_pgd.pdf}
        \caption{PGD}
    \end{subfigure}
        \begin{subfigure}[b]{0.24\columnwidth}
        \includegraphics[width=\columnwidth]{uAdvTFGSMMax_c10_pgd.pdf}
        \caption{uFGSM}
    \end{subfigure}
        \begin{subfigure}[b]{0.24\columnwidth}
        \includegraphics[width=\columnwidth]{uAdvTsgdMax_c10_pgd.pdf}
        \caption{uSGD}
    \end{subfigure}
    \begin{subfigure}[b]{0.24\columnwidth}
        \includegraphics[width=\columnwidth]{figures/FGSM_madry_c10_adam.pdf}
        \caption{FGSM}
    \end{subfigure}
    \begin{subfigure}[b]{0.24\columnwidth}
        \includegraphics[width=\columnwidth]{figures/madry_c10_adam.pdf}
        \caption{PGD}
    \end{subfigure}
        \begin{subfigure}[b]{0.24\columnwidth}
        \includegraphics[width=\columnwidth]{figures/uAdvTFGSMMax_c10_adam.pdf}
        \caption{uFGSM}
    \end{subfigure}
        \begin{subfigure}[b]{0.24\columnwidth}
        \includegraphics[width=\columnwidth]{figures/uAdvTsgdMax_c10_adam.pdf}
        \caption{uSGD}
    \end{subfigure}
    \caption{The universal perturbations made using PGD 
    and ADAM 
    for 4 different robust models trained on CIFAR-10: adversarially trained with FGSM or PGD, and universally adversarially trained with FGSM (uFGSM) or SGD (uSGD). Perturbations were made using 400 iterations.
    The top row perturbations are  made using PGD and the bottom row perturbations are made using ADAM. 
    }
    \label{fig:visual_advTrained_c10_complete}
\end{figure}

We apply PGD (using the sign of the gradient) and ADAM in \cref{alg:adv_univ} to generate universal perturbations for these robust models, and show such perturbations in \cref{fig:visual_advTrained_c10_complete}. 

\section{Universal perturbations for various ImageNet architectures}
\begin{figure*}
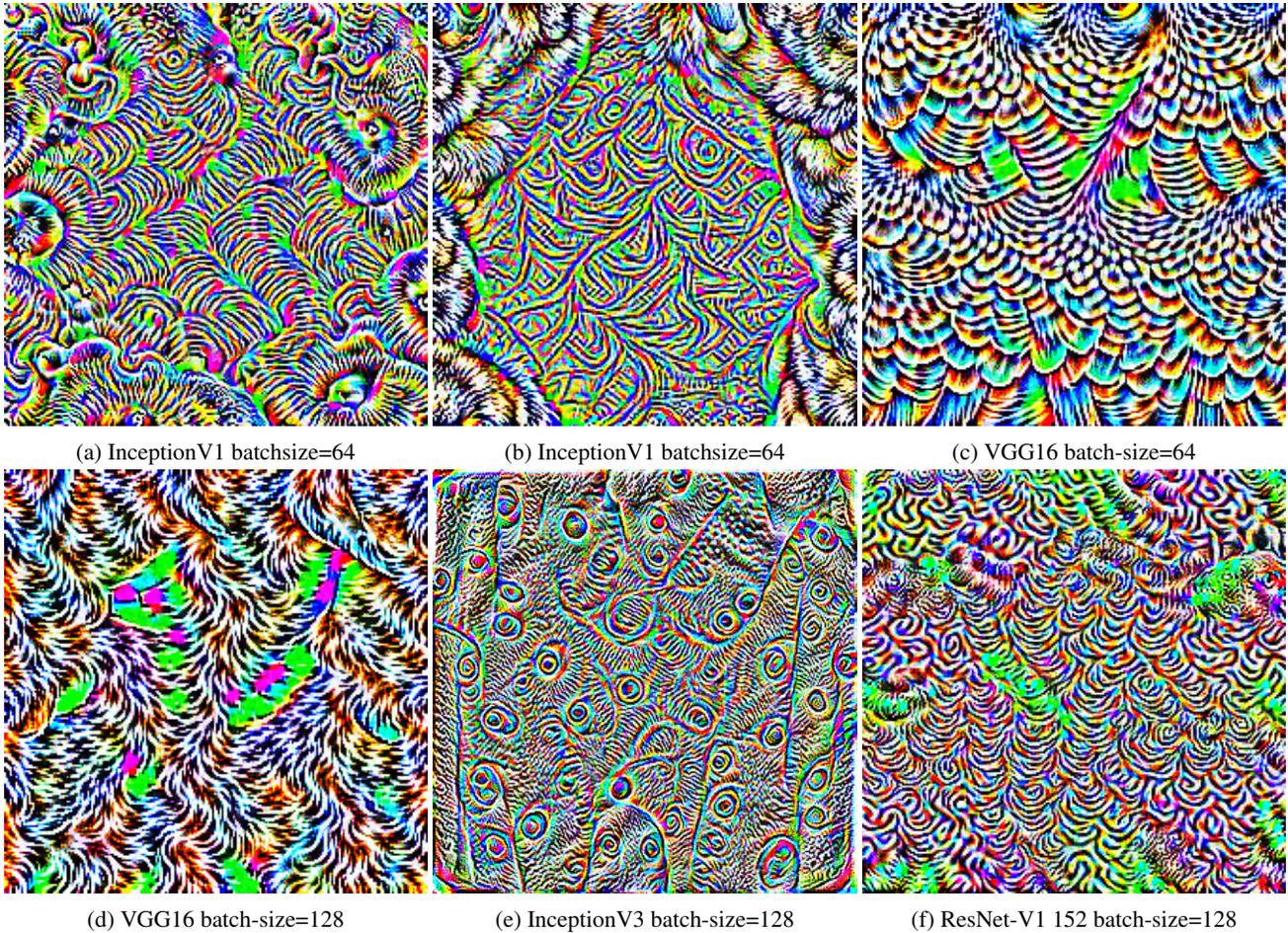

    \centering
    \begin{subfigure}[b]{0.32\linewidth}
        \includegraphics[width=\columnwidth]{figures/inceptionV1_64.pdf}
        \caption{InceptionV1 batchsize=64}
    \end{subfigure}
    \begin{subfigure}[b]{0.32\linewidth}
        \includegraphics[width=\columnwidth]{figures/inceptionV1_128.pdf}
        \caption{InceptionV1 batchsize=64}
    \end{subfigure}
    \begin{subfigure}[b]{0.32\linewidth}
        \includegraphics[width=\columnwidth]{figures/vgg_64.pdf}
        \caption{VGG16 batch-size=64}
    \end{subfigure}
    \begin{subfigure}[b]{0.32\linewidth}
        \includegraphics[width=\columnwidth]{figures/vgg_128.pdf}
        \caption{VGG16 batch-size=128}
    \end{subfigure}
    \begin{subfigure}[b]{0.32\linewidth}
        \includegraphics[width=\columnwidth]{figures/inceptionV3_128.pdf}
        \caption{InceptionV3 batch-size=128}
    \end{subfigure}
    \begin{subfigure}[b]{0.32\linewidth}
        \includegraphics[width=\columnwidth]{figures/resnet152_64.pdf}
        \caption{ResNet-V1 152 batch-size=128}
    \end{subfigure}
    \caption{Universal perturbations generated using our \cref{alg:univ} for different network architectures on ImageNet. 
    Visually, these perturbations which are for naturally trained models are structured.  
    }
    \label{fig:imageNet_noises}
\end{figure*}

In \cref{fig:imageNet_noises}, we plot various universal perturbations found using our ``clipped'' loss attack. Changing the mini-batch size for generating the perturbations, sometimes causes the perturbations for the same architecture to look slightly different.

\section{Comparison with iDeepFool on other datasets}
To ensure that our universal perturbation generation algorithm's performance gain over iDeepFool is not only on ImageNet, we conduct experiments on the WRN 32-10 for CIFAR-10 and LeNet for MNIST. To generate the universal perturbations using both algorithms, we use 5000 training examples and do 10 passes over the data. For CIFAR-10, similar, to the main experiments in the paper, we use $\epsilon=8$. For MNIST, we use $\epsilon=76.5$. The accuracy on the validation images augmented with the universal perturbation are summarized in table~\ref{tab:other_attack_compare}.

\begin{table}
    \centering
    \caption{Top-1 accuracy on CIFAR and MNIST for natural images, and adversarial images with universal perturbations.
    } 
    \begin{tabular}{|c|c|c|c|}
    \hline
    \multicolumn{2}{|c|}{} & CIFAR-10 & MNIST \\ 
    \hline
    Natural & Val & 95.2\% & 99.3\%\\
    \hline
    iDeepFool & Val & 20.0\% & 24.87\%\\
    \hline
    Ours & Val & \bf 13.9\% & \bf 5.7\%\\
    \hline
    \end{tabular}
    \label{tab:other_attack_compare}
\end{table}

\section{Visualizing attacks on robust models}

\begin{figure*}
\centering
    \includegraphics[width=0.98\linewidth]{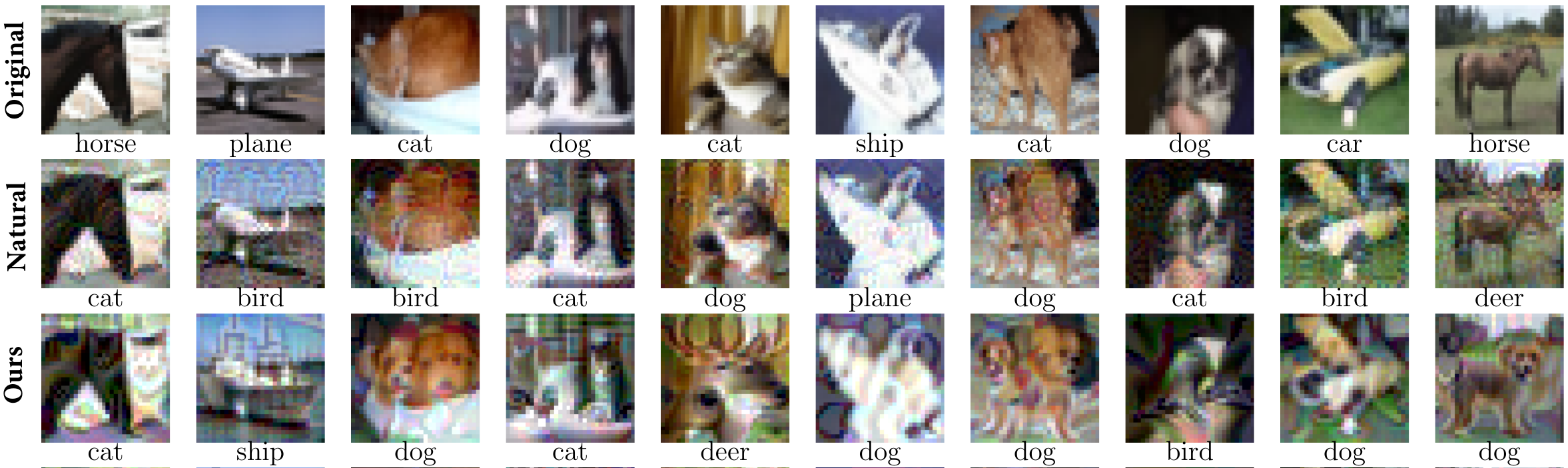}
    \caption{Visualization of the original/natural image and the per-instance adversarial examples generated for the naturally trained and our universally robust model trained with the original version. The examples are the last 10 images of the CIFAR-10 validation set. The adversarial examples have $\epsilon=30$ and are generated using an $l_{\infty}$ 50-step PGD attack with step-size 2. The classifiers' predictions on the examples are printed underneath the images. The large-$\epsilon$ adversarial examples generated for our universal robust model seem to often produce salient characteristics of the targeted class. The predictions of our robust model on the adversarial examples align well with human perception.}
    \label{fig:c10_adv_vis}
\end{figure*}

\cite{tsipras2018robustness} use several visualization techniques to analyze PGD-based robust models and show some unexpected benefits of adversarial robustness.
Similarly, we generate large $\epsilon$ $\ell_\infty$ per-instance adversarial examples using a PGD attack without random initialization. 
Large $\epsilon$ perturbations make the perturbations visible. 
Adversarial examples built in this way for both a natural model and our robust model are illustrated in \cref{fig:c10_adv_vis}. 
Many of the adversarial examples of the natural model look similar to the original image and have a lot of ``random'' noise on the background, while adversarial examples for our robust model produce salient characteristics of another class and align well with human perception. 
The elimination of structured universal perturbations during universal adversarial training seems to have this interesting side-effect that was only recently shown for PGD adversarial training. 

\section{Defense Against per-instance attacks}
Interestingly, while our universal adversarial training alg.~\ref{alg:adv_univ} is for training models that are robust to universal perturbations, we see that when using a strong update rule, the hardened models become robust against $\ell_\infty$ per-instance white-box attacks generated using a 20-step PGD attack. 
While training with the ``normalized'' (FGSM and ADAM) universal perturbation update rules result in models that resist universal perturbations, the FGSM update rule produces models that are more resistant against per-instance attacks compared to the ADAM update rule. 
The accuracy of a universally hardened network against a white-box per-instance PGD attack is 17.21\% for FGSM universal training, and only 2.57\% for ADAM universal training. When compared to FGSM per-instance adversarial training, which has comparable computation cost, the universally robust model is even more robust against per-instance attacks! FGSM per-instance adversarial training achieves 0.00\% accuracy on per-instance adversarial examples built using the same PGD attack setting. More per-instance comparisons are provided in the supplementary material. 

\subsection{Defense against white-box attack}

We compare our universal adversarially trained model's robustness with other hardened models against white-box attacks, where the (robust) models are fully revealed to the attackers. 
We attack the hardened and natural models using universal perturbations and per-instance perturbations (FGSM \cite{goodfellow6572explaining}, R-FGSM \cite{tramer2017ensemble}, and a 20-step $l_{\infty}$-bounded PGD attack with step-size 2 \citep{madry2017towards}). 
We also report the performance of per-instance \textit{adversarially trained} models which are trained with per-instance attacks such as FGSM, R-FGSM and PGD \cite{madry2017towards}. 
We use our original universal adversarial training algorithm to build a robust WideResnet \citep{zagoruyko2016wide} on CIFAR-10 \citep{krizhevsky2009learning}. 
The PGD per-instance adversarial training is done by training on adversarial examples that are built using $7$ steps of PGD following \cite{madry2017towards}, which makes it $4\times$  slower than the non-iterative adversarial training methods such as our universal adversarial training, FGSM, and R-FGSM adversarial training.

We summarize the CIFAR-10 results in \cref{tab:white}. The natural model, is vulnerable to universal and per-instance perturbations. Our robust model achieves best classification (\ie highest robustness) accuracy against universal perturbation attacks. The 20-step PGD attack fools the natural, FGSM robust, and R-FGSM robust models almost every time. Interestingly, our model is relatively resistant to the PGD attack, though not as robust as the PGD-based robust model. This result is particularly interesting when we consider that our method is hardened using universal perturbations. While the computational cost of our method is similar to that of non-iterative per-instance adversarial training methods (FGSM, and RFGSM), our model is considerably more robust against the PGD attack that is known to be the strongest per-instance attack.

{\small
\begin{table}
\centering
\caption{White-box performance of hardened WideResnet models trained on CIFAR-10. We use different attacks to evaluate their robustness. Note that Madry's PGD training is significantly slower than the other training methods. 
} 
\setlength\tabcolsep{3pt}
\begin{tabular}{|c|c|c|c|c|c|}
\hline
\multicolumn{2}{|c|}{} & \multicolumn{4}{c|}{Attack method} \\
\cline{3-6}
\multicolumn{2}{|c|}{} & UnivPert & FGSM & R-FGSM & PGD  \\
\hline
\multirow{4}{*}{\tabincell{c}{(Robust) \\ models \\ trained \\ with}} & Natural & 9.2\% & 13.3\% &
7.3\% & 0.0\%\\
\cline{2-6}
& FGSM & 51.0\% & 95.2\% & 90.2\% & 0.0\%\\
\cline{2-6}
& R-FGSM & 57.0\% & 97.5\% & 96.1\% & 0.0\% \\
\cline{2-6}
& PGD & 86.1\% & 56.2\% & 67.2\% & 45.8\% \\
\cline{2-6}
& Ours & \textbf{91.8\%} & 37.3\% & 48.6\% & \underline{17.2\%}\\
\hline
\end{tabular}%
\label{tab:white}%
\end{table}%
}

\subsection{Transferability and black-box robustness}

 We study the transferability of our robust model in the black-box threat setting, in which we generate adversarial examples based on a source model and use them to attack a target model. 
 We study the transferability of the adversarial 20-step PGD per-instance examples between various models with the WideResnet architecture that are trained on CIFAR-10: natural trained model, FGSM trained robust model, R-FGSM trained robust model, PGD trained robust model \citep{madry2017towards}, and our robust model. 
 
 \begin{table}[ht!]
\centering
\caption{Black-box attack and defense on CIFAR-10. The adversarial examples are generated by PGD. The numbers reported are accuracies.
} 
\setlength\tabcolsep{5pt}
\begin{tabular}{|c|c|c|c|c|c|c|c|}
\hline
& \multicolumn{5}{c|}{Attack source} \\
\cline{2-6}
& Nat. & FGSM & RFGSM & PGD & Ours \\
\hline
Natural & - & 34.1\% & 64.9\% & 77.4\% & 22.0\% \\
\hline
FGSM & 53.9\% & - & 14.1\% & 69.6\%   & 22.7\% \\
\hline
RFGSM & 71.5\% & 16.0\% & - & 	71.7\% & 20.3\%  \\
\hline
PGD & 	84.1\% & 86.3\% & 86.3\% & -  & 76.3\%  \\
\hline
Ours & 90.0\% & 90.8\% & 91.0\% & 70.4\% & -  \\
\hline
\hline
Average & 74.9\% & 56.8\% & 64.1\% & 72.3\% & \textbf{35.4\%}\\
\hline
\end{tabular}
\label{tab:black1}%
\end{table}%

 The results are summarized in \cref{tab:black1}. 
 By examining rows of \cref{tab:black1}, both the PGD-based robust model and our robust model are fairly hardened to black-box attacks made for various source models.
 By examining columns of \cref{tab:black1}, we can compare the transferability of the attacks made for various source models. In this metric, the attacks built for our robust model are the strongest in terms of transferability and can deteriorate the performance of both natural and other robust models. An adversary can enhance her black-box attack by first making her source model universally robust!

\end{document}